Research Article

# Enhancing Digital Health Services: A Machine Learning Approach to Personalized Exercise Goal Setting

Ji Fang[1,2], Vincent CS Lee[2], Hao Ji[3] and Haiyan Wang[1]

## Abstract

**Background:** The utilization of digital health has increased recently, and these services provide extensive guidance to encourage users to exercise frequently by setting daily exercise goals to promote a healthy lifestyle. These comprehensive guides evolved from the consideration of various personalized behavioral factors. Nevertheless, existing approaches frequently neglect the users' dynamic behavior and the changing in their health conditions.
**Objective:** This study aims to fill this gap by developing a machine learning algorithm that dynamically updates auto-suggestion exercise goals using retrospective data and realistic behavior trajectory.
**Methods:** We conducted a methodological study by designing a deep reinforcement learning algorithm to evaluate exercise performance, considering fitness-fatigue effects. The deep reinforcement learning algorithm combines deep learning techniques to analyse time series data and infer user's exercise behavior. In addition, we use the asynchronous advantage actor-critic algorithm for reinforcement learning to determine the optimal exercise intensity through exploration and exploitation. The personalized exercise data and biometric data used in this study were collected from publicly available datasets, encompassing walking, sports logs, and running.
**Results:** In our study, we conducted The statistical analyses/inferential tests to compare the effectiveness of machine learning approach in exercise goal setting across different exercise goal setting strategies. The 95% confidence intervals demonstrated the robustness of these findings, emphasizing the superior outcomes of the Machine Learning approach.
**Conclusions:** Our study demonstrates the adaptability of machine learning algorithm to users' exercise preferences and behaviors in exercise goal setting, emphasizing the substantial influence of goal design on service effectiveness.

[1]School of Economics and Management, Southeast University, Nanjing, China
[2]Department of Data Science and Artificial Intelligence, Monash University, Melbourne, Australia
[3]Hangzhou Medical College, Hangzhou, China
**Corresponding author:**
Haiyan Wang, School of Economics and Management, Southeast University, Nanjing, 211189, China. Email: hywang@seu.edu.cn

# Introduction

Improving the efficacy of digital health services requires the combination of Internet of Things (IoT) technology and machine learning methodology.[1,2] IoT devices with embedded sensors enable the automatic measurement and remote monitoring of human physical activity during exercise sessions. This capability facilitates the collection of comprehensive data regarding the health-related activity patterns of users, particularly in relation to their daily activities.[3] In this paper, an exercise refers to a subset of physical activity that is planned and executed with a specific goal and has as its ultimate or intermediate goal of improvement or maintenance of exercise.[4,5] The digital health service, through the utilization of real-time situation triggered reminders, pushes, and notifications, can assist the user in improving the effectiveness of daily exercise.[6] Building upon the accurate cognitive health assessment provided by IoT devices, machine learning technologies in the digital health service provide personalized exercise guidance tailored to the physical needs of users,[7] and offer achievement rewards by incorporating external incentives (e.g., virtual badges) into the exercise process.[8] Recent research has emphasized the potential of machine learning in strengthening health intervention services.[9-10] These studies emphasize the ongoing development and significance of integrating advanced data-driven methodologies into digital health services. Moreover, machine learning applications have made notable advancements in various health fields, such as epilepsy diagnostics[11], and cancer prognosis.[12] The results of these studies correspond with previous several systematic reviews and empirical studies that have demonstrated the efficacy of digital health services in exercise guidance.[13-15]

Despite the considerable potential of digital health services in promoting exercise regardless of time and location constraints, there are persistent challenges in evaluating and improving exercise guidance in digital health service. Several factors contribute to the inefficiency of the exercise guidance in the digital health service. First, the majority of popular digital health services offer non-personalized, predetermined exercise plans. There is mounting evidence that such fixed daily exercise plans may demotivate users from exercising.[16] Secondly, the exercise goals can cause difficulties if they are excessively difficult or too simple, resulting in user overload or underload. We adapt the Fitness-Fatigue theory as a framework for comprehending an individual's performance and response to exercise workload. The Fitness-Fatigue theory, as conceptualized by Calvert et al.,[17] provides a foundational framework for understanding the impact of training on exercise performance combining the fundamental principles of exercise science and training theory. It comes with the assumption that every exercise session generates two opposing responses: "Fitness", which signifies the underlying adaptations contributing to long-term performance improvements, and "Fatigue", which relates to transitory reductions in exercise performance. Consequently, "Fitness" and "Fatigue" provide values associated with the intensity of how much exercise has been performed recently. In this context, "Fatigue" can manifest in both physical and mental aspects and typically refers to the immediate effects of exercise, including muscle fatigue, decreased coordination, and a feeling of exhaustion.[17] Thirdly, the evaluation of the performance of the digital health service is dependent on the user's adherence to the exercise plans.[18] The user may struggle with self-control problems, resulting in failure to complete the daily plans.[19] Moreover, the time-consuming nature of the digital health service, the long duration of exercise programs, and the unpredictability of future health outcomes can all contribute to user attrition from the digital health service. This is because the users may not experience immediate benefits from their exercise efforts and therefore lack the motivation to continue. In summary, the heterogeneity of users' physical characteristics and exercise preferences causes the complexity of exercise goal setting. The path differences between planned and actual user behavior can lead to unpredictable digital health service performance.

As a solution, this paper proposes a machine learning framework for the exercise goal setting problem in digital health service that incorporates Fitness-Fatigue theory to model the exercise process and provides exercise guidance. The machine learning framework that considers the exercise preferences of users is essential to enhancing their engagement in exercise. Constructing a real-time interaction model between the digital health service system and the user enables dynamic feedback. If the user's behavior varied from the exercise plans, the proposed machine learning algorithm could dynamically modify the exercise goal setting strategy during each interaction. Tailoring the exercise plan to meet users' specific needs increases their sense of accomplishment and ensures that they are challenged without overexerting themselves, resulting in a positive experience that maintains their exercise motivation. The experiments were conducted to test how our machine learning framework would have performed with various user responses in a digital health. In our study, we hypothesized that the efficacy of our machine learning approach for exercise goal setting problem would be superior to that of fixed exercise goal setting strategies. In addition, we hypothesized that it would produce better results than in without exercise goal setting service. The results of experiments revealed that our machine learning approach outperformed other exercise goal setting strategies. In addition, the findings indicated that our exercise goal setting strategy outperformed other non-adaptive strategies by 4.31 percent. The sensitive analyses provided insights into the incentive effects of exercise goal setting in digital health services.

The overall objective of our work was to propose a machine learning methodology for personalized exercise goal setting in digital health that focused on user exercise behavior factors, utilizing exercise data and fitness profiles to provide optimized exercise recommendations. To do this, we presented a knowledge-based digital health system that utilized the exercise performance evaluation model to select exercise goals while considering the effects of fitness and fatigue. Furthermore, we developed a machine learning algorithm by constructing the real-time interaction model to dynamically control the exercise goal setting for each user, adapting to changes in the user's exercise preferences, adherence behavior, and uncertain exercise performance. To validate the proposed machine learning algorithm, we conducted experiments using actual data and simulated various user responses to the digital health service. Our findings contributed to the development of digital health through exercise and the improvement of healthy lifestyles.

# Methods

## Model Framework

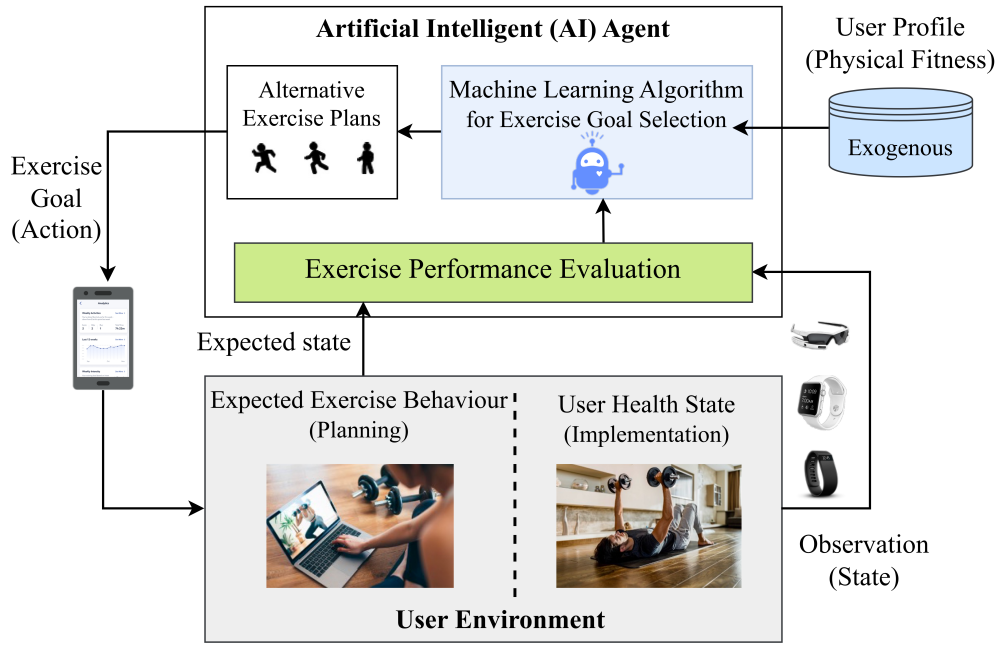

**Figure 1.** The exercise goal setting model in digital health in an interaction environment.

The exercise goal setting system is designed within an interactive environment (Figure 1). The Artificial Intelligence (AI) agent utilizes IoT devices to monitor the user's exercise behavior at the beginning of each interaction. The AI agent then selects a short-term exercise goal from a pool of prospective plans, ensuring that it meets the user's fitness needs. The selection mechanism of the AI agent is the evaluation of the user's exercise performance, which provides valuable feedback for the exercise process optimization flow. This optimization procedure simulates realistic behavior trajectories based on historical exercise data. After receiving the exercise suggestion information via fitness apps, the user tries to exercise. The AI agent receives observations of the user's actual exercise behavior and monitors any changes to their health status. If the user deviates from the prescribed exercise plan, the AI agent will adjust its exercise goal setting strategy accordingly.

In this study, we focus on three primary aspects to improve the performance of the digital health service in exercise goal setting for each user. First, we recognize that altering the exercise preferences of the user can affect their exercise behavior, thereby influencing the intervention effect of exercise goal setting in digital health. Second, we suppose that the digital health service can only decide the exercise suggestions, not the user's actual behavior, based on realistic consideration. Third, we consider that exercise plans with the same intensity level can exert various levels of stress on different users, highlighting the need for personalized exercise goal setting that accounts for the user's physiological characteristics. This study seeks to optimize the personalized exercise process through real-time monitoring, feedback, and adaptive goal setting to increase users' engagement, adherence, and ultimately exercise performance.

## Practical Considerations for the Model

### *Expected Exercise Behavior*
In this study, we assume that the user can strictly adhere to the exercise plan, which prevents overestimation or underestimation of the user's realization of exercise goals. This assumption provides a reliable estimate of the user's exercise performance. However, we noticed that deviations from the exercise plan could occur. To address this, the machine learning algorithm is designed to adjust the exercise suggestions in case the user does not finish the exercise plan. The AI agent utilizes self-tracking devices to measure the actual exercise intensity level. Based on this information, the AI agent dynamically adjusts the estimated exercise performance and updates the fitness guidance for the remaining service period. This real-time communication between the user and the AI agent enables ongoing decision-making for personalized exercise goal setting.

### *Exercise Performance Evaluation*
In this paper, it is supposed that the evaluation of exercise performance consists of two components: the performance resulting from exercise effort, and the intervention effect of exercise goal setting in digital health service.[20,21] For the measurement of exercise performance, we drew insights that exercise performance incorporates the effects of each workout on the "stock of fitness" and the "stock of fatigue".[22] These stocks are measured with respect to the user's baseline exercise level and exhibit exponential decay over time intervals. For the measurement of the intervention effect of digital health, we consider the influence of exercise goals on users' mental achievement, building upon the work of Aswan et al.[23] They examined and quantified the impact of achieving and failing to complete exercise goals, providing the foundations for the effects of goal attainment on user motivation.

Within the field of health management, exercise skills are divided into two stages:[22] skill acquisition and skill retention. Repeated exercise has a compounding effect on memories via a "re-consolidation" process, whereas exercise cannot have a compounding effect until early skill acquisition occurs. As a result, the additive model is considered to be more appropriate for the stage of exercise skill acquisition, while the multiplicative model is considered to fit the exercise skill retention stage. In this paper, we assume that the user's exercise skill stage remains fixed for one service period. The user is required to choose their exercise type, either as a beginner or an advanced user.

The evaluation framework in this paper provided a comprehensive understanding of exercise performance by incorporating the effects of exercise self-effort, the stocks of fitness and fatigue, exercise goal attainment, and the different stages of exercise skill levels. This framework enables the AI agent to assess the impact of exercise goal setting on the user's exercise outcomes and guide the decision-making process of exercise goal setting within the digital health service. The notations and the performance evaluation function are presented in the Section A and Section B in the Supplementary material.

*User Profile and Health State*
The user's physical fitness information is derived from exogenous data, specifically through the analysis of the user's profile history data. Within the user profile, we define $\chi = (\alpha, \beta, \lambda, \mu, \delta, k_f, k_g, m, l)$ as the user type. The parameters $\alpha, \beta$ ranging from 0 to 1, represent the decay rates of fitness and fatigue, respectively. These rates reflect the natural processes of forgetting (for fitness) or recovering (for fatigue) over time. In addition, $\lambda, \mu$ describe the nonlinearities in the responses of fitness and fatigue to the exercise effect. Specifically, $\lambda$ accounts for the concave effect on fitness, meaning that as exercise intensity increases, the rate of fitness improvement diminishes. On the other hand, $\mu$ represents the convex effect on fatigue, indicating that fatigue accumulates at an accelerating rate as exercise intensity increases. Moreover, the parameter $0 < \delta \leq 1$, quantifies the decay in the user's base fitness level, considering the temporal effects on base fitness level. The parameters $k_f, k_g$, both greater than 0, represent the marginal utility of fitness and fatigue levels, respectively. They capture the benefits associated with improvements in fitness and the negative effects associated with fatigue. The $m$ is a nonnegative number, captures the increase in utility experienced if the exercise plan is achieved. Based on reasonable assumptions, it is supposed that regardless of how much exercise intensity level a user achieves, the increase in utility for goal achievement is the same. The $l$ is a nonnegative number, represents the impact of failing the exercise plan. The greater the degree of the unfinished exercise, the greater the decrease in utility experienced due to frustration.

The user's health state, denoted as $s_t = (e_t, b_t, f_t, g_t)$, is a collection of state variables obtained through IoT devices, recording the dynamic exercise process. The subscript $t$ refers to the $t$-th decision epoch of the user's health state, where $t = 0, 1, \ldots, T$. $e_t$ is a nonnegative number, quantifies the intensity of the user's exercise workout at time $t$. It captures the level of effort put into the exercise workout in the decision epoch. The $b_t$ is a nonnegative, represents the base level of past exercise. The $f_t, g_t$ represent the fitness and the fatigue at time instance $t$ respectively. The fitness refers to the positive effects of exercise such as increased strength, endurance. The fatigue refers to the temporary decrease in exercise performance in both mental fatigue and physical fatigue, such as muscle fatigue, a feeling of exhaustion. Our research focuses predominantly on estimating the values of fitness and fatigue using established mathematical formulas [22].

*Exercise Goal Selection*
The AI agent seeks a decision-making strategy to maximize exercise performance during a finite service period. Figure 2 depicts the sequence of the interaction process within a single decision epoch. At the beginning of the decision epoch $t$, the AI agent chooses an exercise plan with an exercise goal $a_t$. The agent will then evaluate the exercise performance with the assumption that the user can complete the exercise plan. Exercise performance evaluation considers the exogenous information associated with the personal profile data (i.e., in the skill acquisition stage or in the skill retention stage, the physical fitness data). The exercise performance is also related to the user's former health state $s_{t-1}$. The user tries to carry out the exercise plan, and the user's real exercise intensity level $e_t$ is observed by the AI agent. The user's health state $s_t$ is updated in the digital health service. We define the exercise goal setting problem using the Markov Decision Process and culminating in the presentation of the optimal personalized goal setting strategy expression (Section C in the Supplementary Material).

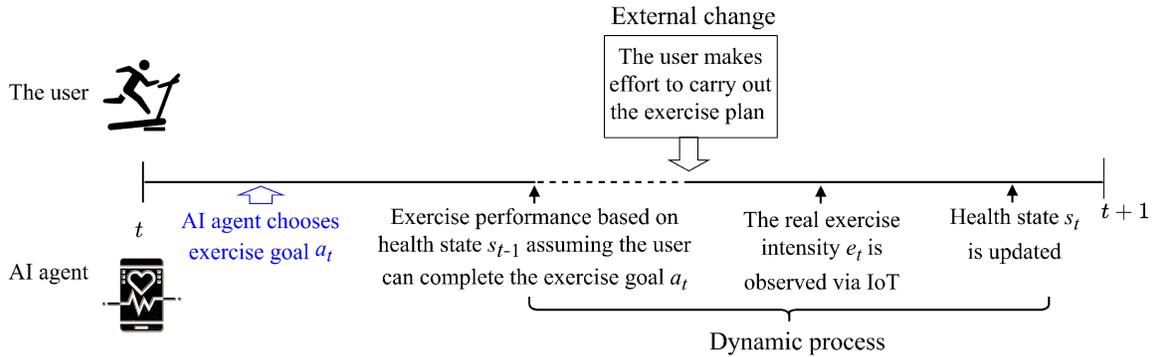

**Figure 2.** The sequential decision-making process in an interaction between a user and a system.

## The Proposed Solution Using Machine Learning

To address the dynamic, uncertain, and uncontrollable nature of user exercise behavior in digital health services, we employ a machine learning approach that learns from user experiences through the sampling method. Unlike traditional optimization methods, our machine learning algorithm does not rely on assumptions about the user's exercise behavior, as it considers the unpredictable impact of external influences. In our algorithm, we gather the interaction data to extract sequences of states, actions, and rewards. Furthermore, we enhance the machine learning method by constructing a hybrid neural network. This network incorporates a Long Short-Term Memory (LSTM) neural network, allowing it to capture temporal information from the user's exercise history. By combining this with a fully connected neural network, we improve the model's ability to fit the data effectively.

In this study, the exercise goal setting strategy is trained using a combination of the generalization method (actor network) and the parameterized policy method (critic network).[24] The machine learning algorithm trains our hybrid neural network using two major techniques: asynchronous training agents and the advantage function in the policy gradient. Figure 3 illustrates the basic structure of AI agents.

The details of the machine learning approach and the algorithm are presented in the Section D in the supplementary Material.

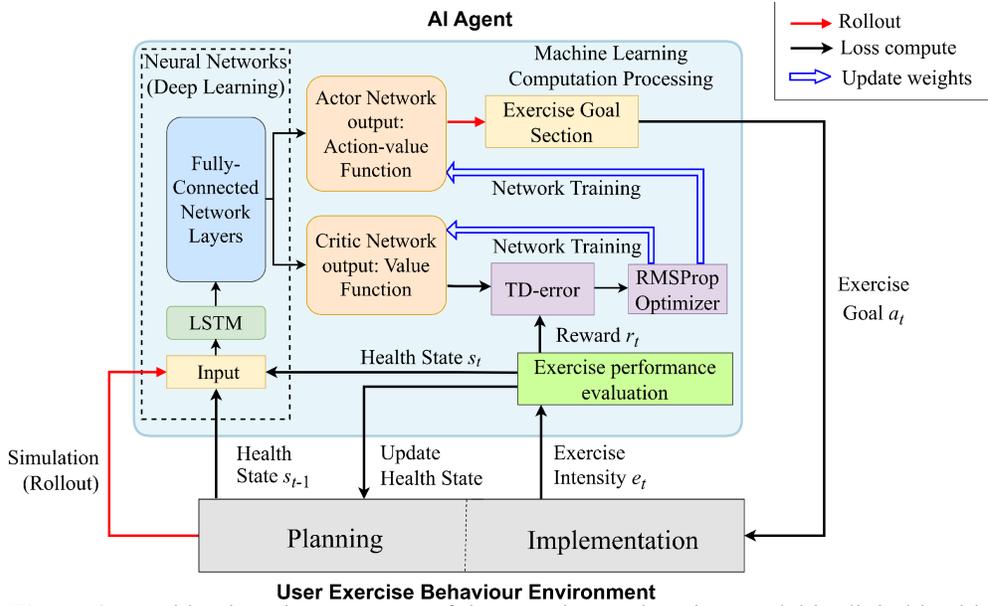

**Figure 3.** Machine learning structure of the exercise goal setting model in digital health service.

The proposed machine learning algorithm combines the concepts of actor-critic methods with parallelization and neural networks. The computational complexity of the proposed algorithm in each episode can be broken down into several components: (1) Policy evaluation: The policy function is computed through the actor network requiring a computation time $O(T_a)$, which is determined by the network structure (e.g., the number of hidden layers, the number of hidden units, the type of neural cell).[25] (2) Value evaluation: The value function is computed through the critic network, requiring a computation time $O(T_c)$ the same as the actor network. (3) Exercise performance calculation: Computing the reward function at each time step takes $O(1)$ time. (4) TD-error calculation: The TD-error calculation requires $O(1)$ time as it involves a simple subtraction of the value function from the reward. (5) Parameter update: Updating the parameters of the policy and value functions at each time step. The overall time complexity of the machine learning algorithm can be described as $O(T_{max}(T_a + T_c + 2)/N_l)$ where $T_{max}$ is the number of time steps. $N_l$ is the number of agents. The computational time required by the proposed algorithm grows linearly with the number of time steps $T_{max}$, but is influenced by the size of neural networks, and the number of agents.

## Modeling User Behavior in the Digital Health Service

In order to create the simulation environment for the machine learning approach, we conducted a thorough review of existing scientific literature that focused on increasing exercise levels through digital health interventions.[26,27] According to the findings of this literature, the changing trends in users' exercise levels with digital health services were categorized into four distinct types. Several studies have indicated that digital health interventions are not effective in increasing exercise.[28-30] For instance, Kirwan et al. conducted a 90-day study in which they compared an intervention group utilizing a website-delivered exercise program with a matched group.[28] However, they found no significant relationship between the perceived usability and usefulness of the digital health service and exercise levels. On the other hand, the majority of interventions reported significant increases in exercise levels through the use of digital health services.[27,31] For this study, we considered two degrees of exercise level increase. The design of the fourth environment was based on the observations from a sequential multiple assignment randomized trial and observed both decreasing and increasing trends in the number of steps per day throughout the intervention.[32]

To evaluate the practicality of the proposed machine learning approach in the digital health service, we conducted a series of experiments, as presented in **Table 1**. These experiments aimed to test different types of user behavior changes by constructing the environment based on the diverse responses of users to the digital health service.

**Table 1.** User exercises change behavior and experiment settings.

|  | **Behavior Changing Tread** | **References** | **Environment emulator** |
|---|---|---|---|
| **E1** | No effect | [28-30] | Remain stable for 12 weeks |
| **E2** | Sightly increase | [31] | Increase by 40% after 6 weeks |
| **E3** | Highly increase | [27] | A two-fold increase after 6 weeks |
| **E4** | A down and an increase | [32] | Declines by 20% for 6 weeks, increase by 60% after 6 weeks |

## User Study and Data Collection

The digital health data used in the study involved three groups of users, with a brief description provided in **Table 2**. The corresponding metrics and their units in the experiments are presented in Table 3. Data for the first study (G1) was generated based on existing literature, with randomized individual-specific parameters and randomly generated daily exercise intensity.[28]

In the second study (G2), data was sourced from an open-source data set by Thambawita et al., collected from 16 individuals over five months, except for five participants with missing data.[33] Participants' perceived exertion level was chosen as the exercise intensity. The PMDATA was pre-processed through the integration of traditional lifelogging data with sports-activity data, enabling the development of health-related analyses. This dataset combined input from Fitbit Versa 2 smartwatch wristbands, the PMSys sports logging smartphone application, and Google forms. We evaluated the data quality based on the data consistency, validity, and accuracy. The consistency of the dataset was maintained by covering the reporting period from November 2019 to March 2020. The dataset was collected from a sample size of 16 participants. The diversity of training and exercise background among the participants improved the validity of the dataset. Some are active athletes, some previous athletes, and some rarely exercise at all. In terms of data accuracy, the sports-activity data was automatically recorded by Fitbit Versa 2 smartwatch wristbands. A limitation of this dataset is the presence of noise resulting from its time-series data, which requires addressing missing data and identifying outliers. As a result of this constraint, we opted to exclude five participants from the dataset who had missing data. Another constraint of the dataset involves addressing technical obstacles related to the integration of heterogeneous data sources. We selectively utilised specific indicators relevant to our research assumptions rather than using the whole dataset. The assessment of exercise intensity in our study was based on the subjective perception of exertion levels reported by the participants.

For the third study (G3), our research team collected data from 13 adult participants over a period of 12 weeks. During the data collection process, we endeavored to ensure the data quality regarding consistency, relevance, timeliness, and data validity. This study was conducted as a retrospective study, no human or animal trials were involved. The primary objective of this study was to collect retrospective log data to simulate the real world user exercise situation. There were no specific exclusion criteria, and all willing adults in our research group were included in the study. The data was collected through the Fitbit Connection Project managed by Open Humans Foundation protecting user anonymized and deidentified, and informed consent. The participants were equipped with Fitbit Charge 3 wearable devices to measure exercise intensity during their daily runs. The collected data was pre-processed based on professional knowledge of running science. For details on the data analysis in running science.[34] Running intensity was estimated using Training Impulses from the pre-processed data, and exercise performance was evaluated using $VO_2Max$ User-specific parameters, which were then estimated using the nonlinear least-squares method in Python, minimizing the residual sum between the estimated performance and the observed $VO_2Max$ performance data. One potential constraint of this dataset is the limited sample size of 13 adult participants, which may restrict the generalisability of the findings to a larger population. Additionally, it is important to consider the potential assumptions and limitations associated with the selected estimation methods for evaluating running intensity and exercise performance, since these factors can impact the results obtained.

**Table 2.** Summary statistics of the data set.

|    | Data set | Participants | Exercise type | Indicator selection | Data Source |
|----|----------|--------------|---------------|---------------------|-------------|
| **G1** | Simulation | 50 participants (26 males, 24 females) Age range: 17-64 year | Walking | Walking steps (mean$\pm SD$) 6274$\pm$2106 | [28] |
| **G2** | Open source | 11 participants (8 males, 3 females) Age range: 26-60 year | Exercise Log | Perceived exertion (mean$\pm SD$) 2.43$\pm$3.02 | [33] |
| **G3** | Real world | 13 participants (7 males, 6 females) Age range: 25–33 year | Running | Training Impulses (mean$\pm SD$) 21.35$\pm$14.15 | Collected from Fitbit |

**Table 3.** The units of the corresponding metrics in the experiments

| Measure | Definition | Metric | Data Source |
|---------|------------|--------|-------------|
| **Exercise Intensity** | Measure the the effort put in exercise | Walking steps | Random generated (G1) |
| | | The perceived exertion | Scrpe.csv in the PMSys (G2) |
| | | Training IMPulses | Collected from Fitbit (G3) |
| **User Profile** | Exogenous information about the user | $\chi = (\alpha, \beta, \lambda, \mu, \delta, k_f, k_g, m, l)$ | Random generated (G1, G2) |
| | | Maximal oxygen consumption ($VO_2Max$) as the exercise performance, using nolinear least-squares to estimate user profile | $VO_2Max$ is estimated from Fitbit data (G3) |

# Results

To evaluate the performance of the proposed machine learning algorithms and compare them against the performance without any digital health intervention service, we tested three experiment objectives (G1, G2, and G3) with different assumptions that the four types of user's exercise behavior change (Environment E1, E2, E3, and E4). Each simulation was run 100 times for statistical purposes. We observed that each simulation takes less than 370 seconds and can be executed in 3 seconds during testing. The execution time was close to real time. We normalized the exercise intensity with the historical value of the highest exercise intensity and the lowest exercise intensity. We grouped the exercise plans with different intensity levels into five types, where the value 0 presents without any digital health service, values 0.1, 0.2 belong to weak exercise intensity, values 0.3, 0.4, and 0.5 belong to slightly weak exercise intensity, values 0.6, 0.7, and 0.8 belong to slightly strong exercise intensity, and values 0.9, 1 belong to strong exercise intensity.

For demonstration purposes, Figure 4 displays the daily rewards and the clustering of daily exercise goal setting during the 30-day service period. It can be observed that the machine learning approach can adapt to the user's exercise behavior because it causes larger fluctuations and performs better with the daily exercise reward than without the digital health service. Besides, it can be observed that when the user is at a high level of exercise intensity, the machine learning approach may cause slightly better performance, and it is more inclined to choose high intensity exercise plans than in other environments.

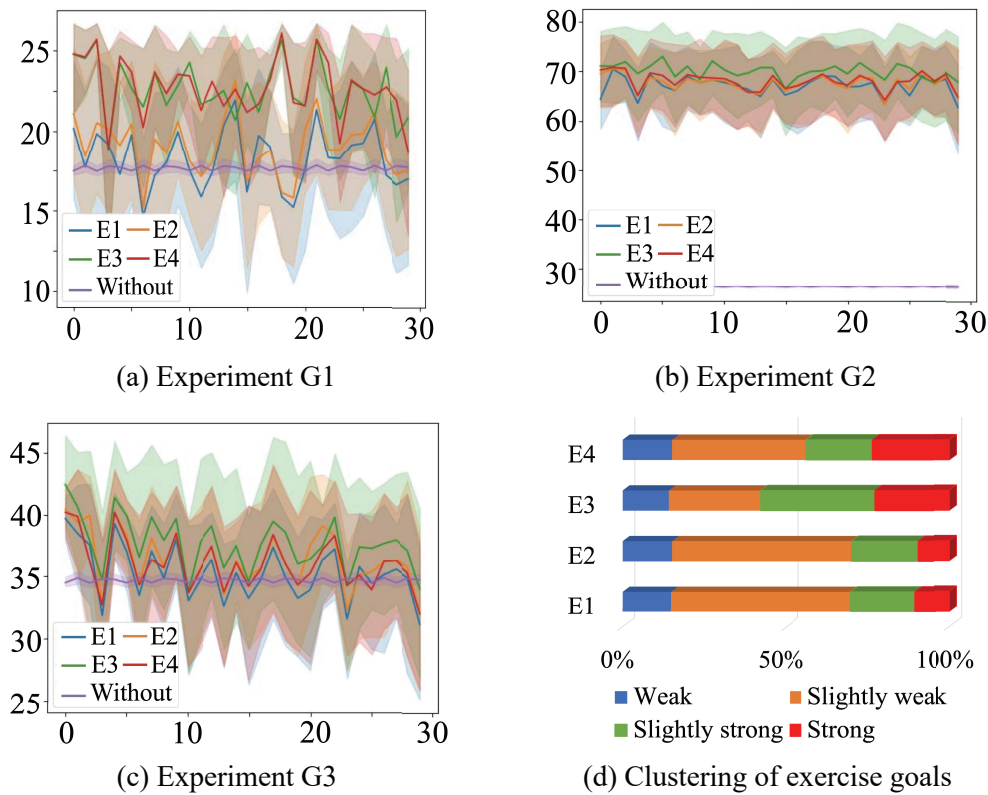

**Figure 4.** Visualization of the experiment result for the machine learning approach on digital health data in the skill acquisition stage (taking the 30-day service time as the axis).

In addition, we conducted comparison experiments between our machine learning approach and the existing fixed exercise goal setting policies. We considered five exercise goals with different exercise intensity levels (Table 4).

**Table 4.** Total rewards improvement of the machine learning approach over other strategies.

(a) Experiments in G1 for skill acquisition stage

| Strategies | E1 | E2 | E3 | E4 |
|---|---|---|---|---|
| Machine learning approach (Mean) | 535.12 | 539.23 | 676.38 | 675.25 |
| Weak intensity fixed strategy | 31.28% | 5.82% | 4.30% | **4.13%** |
| Slightly weak intensity fixed strategy | 15.05% | 15.33% | 4.60% | 13.80% |
| Slightly strong intensity fixed strategy | 59.86% | 57.32% | 20.83% | 38.66% |
| Strong intensity fixed strategy | 162.78% | 155.46% | 47.22% | 83.26% |
| Without recommendation strategy | 5.74% | 6.55% | 33.65% | 33.43% |

(b) Experiments in G2 for skill acquisition stage

| Strategies | E1 | E2 | E3 | E4 |
|---|---|---|---|---|
| Machine learning approach (Mean) | 2149.39 | 2184.89 | 2242.52 | 2179.61 |
| Weak intensity fixed strategy | 6.44% | 4.56% | 6.47% | **4.31%** |
| Slightly weak intensity fixed strategy | 8.99% | 4.45% | 6.71% | 4.90% |
| Slightly strong intensity fixed strategy | 15.28% | 12.08% | 12.13% | 14.28% |
| Strong intensity fixed strategy | 26.32% | 21.75% | 18.13% | 25.08% |
| Without recommendation strategy | 202.64% | 207.63% | 215.7% | 206.89% |

(c) Experiments in G3 for skill acquisition stage

| Strategies | E1 | E2 | E3 | E4 |
|---|---|---|---|---|
| Machine learning approach (Mean) | 1253.81 | 1320.89 | 1374.30 | 1325.33 |
| Weak intensity fixed strategy | 13.59% | 5.88% | 8.46% | **5.04%** |
| Slightly weak intensity fixed strategy | 20.74% | 20.78% | 22.07% | 24.89% |
| Slightly strong intensity fixed strategy | 65.43% | 68.65% | 58.26% | 69.57% |
| Strong intensity fixed strategy | 102.48% | 104.81% | 88.08% | 106.57% |
| Without recommendation strategy | 6.92% | 14.44% | 16.95% | 13.03% |

(d) Experiments in G1 for skill retention stage

| Strategies | E1 | E2 | E3 | E4 |
|---|---|---|---|---|
| Machine learning approach (Mean) | 1099.30 | 1206.48 | 1350.24 | 1278.54 |
| Weak intensity fixed strategy | **4.94%** | 7.40% | 18.83% | 14.46% |
| Slightly weak intensity fixed strategy | 24.13% | 20.39% | 31.37% | 102.61% |
| Slightly strong intensity fixed strategy | 110.58% | 75.07% | 64.41% | 119.89% |
| Strong intensity fixed strategy | 150.63% | 142.31% | 99.54% | 185.56% |
| Without recommendation strategy | 41.14% | 54.90% | 73.32% | 64.16% |

(e) Experiments in G2 for skill retention stage

| Strategies | E1 | E2 | E3 | E4 |
|---|---|---|---|---|
| Machine learning approach (Mean) | 1641.85 | 1734.88 | 1782.40 | 1170.82 |
| Weak intensity fixed strategy | 13.28% | 30.23% | 20.78% | **12.90%** |
| Slightly weak intensity fixed strategy | 30.79% | 80.02% | 45.82% | 19.24% |
| Slightly strong intensity fixed strategy | 107.94% | 102.90% | 47.90% | 49.24% |
| Strong intensity fixed strategy | 127.53% | 123.95% | 63.44% | 71.10% |
| Without recommendation strategy | 26.76% | 33.94% | 36.72% | 36.72% |

(Continue of Table 4)

(f) Experiments in G3 for skill retention stage

| Strategies | E1 | E2 | E3 | E4 |
|---|---|---|---|---|
| Machine learning approach (Mean) | 1235.40 | 1128.96 | 1325.14 | 1263.84 |
| Weak intensity fixed strategy | 11.90% | 16.16% | 8.44% | **6.53%** |
| Slightly weak intensity fixed strategy | 19.17% | 79.45% | 40.20% | 82.46% |
| Slightly strong intensity fixed strategy | 38.89% | 101.33% | 73.55% | 104.37% |
| Strong intensity fixed strategy | 75.57% | 98.91% | 78.44% | 107.31% |
| Without recommendation strategy | 5.51% | 9.40% | 13.18% | 7.94% |

(End of Table 4)

The complete experiment results table is presented in the Section E in the supplementary Material. To visually illustrate the experimental results, we present the results for G3 in the skill acquisition stage as an example in Figure 5. Interestingly, in some cases, the fixed policies with low exercise intensity outperformed our proposed strategy. This outcome can be attributed to situations where users are physically inactive and more likely to adhere to low-intensity exercise plans. Furthermore, encountering obstacles and difficulties during high-intensity exercise plans may discourage users.

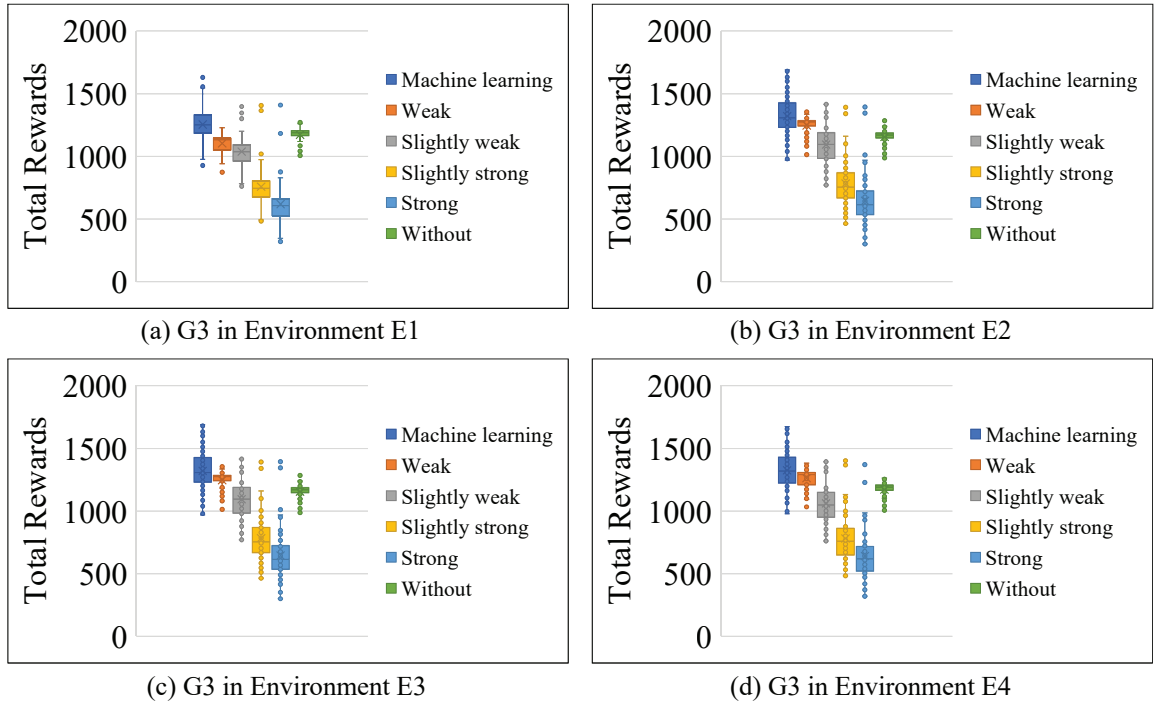

**Figure 5.** Comparisons with fixed strategies testing in G3 for skill acquisition stage.

## Statistical Analysis

To assess the impact of the proposed machine learning approach on user's exercise performance improvement, we conducted an analysis of variance (ANOVA). We formulated and tested the hypotheses to investigate the effectiveness of our machine learning approach in enhancing exercise performance. Our null hypothesis ($H_0$) posits that there are no significant differences in the value of exercise performance between machine learning approach and fixed exercise goal setting strategies, including fixed weak exercise intensity, slightly weak exercise intensity, slightly strong exercise intensity, and strong exercise intensity, and without exercise goal setting service. On the other hand, the alternative hypothesis ($H_a$) posits that there are significant differences in the value of exercise performance between machine learning approach and other strategies. The significance level $a$ was set as 0.05 in the experiments. The descriptive results shown in Table 5. In Table 6, we shown the multiple comparisons between machine learning approach and other strategies.

**Table 5.** Descriptive statistics.

|  | N | Mean | Std. Deviation | Std. Error | 95% Confidence Interval Lower Bound | 95% Confidence Interval Upper Bound | Minimum | Maximum |
|---|---|---|---|---|---|---|---|---|
| Machine Learning | 2400 | 1392.77 | 498.541 | 10.176 | 1372.82 | 1412.73 | 472 | 2368 |
| Weak Intensity | 2400 | 1260.91 | 461.356 | 9.417 | 1242.45 | 1279.38 | 402 | 2181 |
| Slightly Weak | 2400 | 1108.14 | 497.626 | 10.158 | 1088.22 | 1128.06 | 440 | 2184 |
| Slightly Strong | 2400 | 924.76 | 502.474 | 10.257 | 904.65 | 944.87 | 274 | 2108 |
| Strong Intensity | 2400 | 808.26 | 490.849 | 10.019 | 788.62 | 827.91 | 140 | 2032 |
| Without Service | 2400 | 941.66 | 284.716 | 5.812 | 930.26 | 953.06 | 526 | 1310 |
| Total | 14400 | 1072.75 | 505.171 | 4.210 | 1064.50 | 1081.00 | 140 | 2368 |

**Table 6.** Inferential Statistics results of multiple comparisons between machine learning approach and other strategies.

| (I) | (J) | Mean Difference (I-J) | Std. Error | p-value | 95% Confidence Interval Lower Bound | 95% Confidence Interval Upper Bound |
|---|---|---|---|---|---|---|
| Machine Learning | Weak Intensity | 131.858* | 13.352 | 0.000 | 92.66 | 171.05 |
|  | Slightly Weak | 284.635* | 13.352 | 0.000 | 245.44 | 323.83 |
|  | Slightly Strong | 468.012* | 13.352 | 0.000 | 428.82 | 507.21 |
|  | Strong Intensity | 584.509* | 13.352 | 0.000 | 545.31 | 623.70 |
|  | Without Service | 451.111* | 13.352 | 0.000 | 411.92 | 490.31 |

*. The mean difference is significant at the 0.05 level.

The results of our multiple comparisons, as presented in Table 6, indicated that our machine learning approach yielded statistically significant mean differences when compared to all the mentioned fixed exercise goal setting strategies. The significant p-values ($p < 0.001$) strongly supported the rejection of the null hypothesis in favour of the alternative hypothesis. These findings confirm that our machine learning algorithm leads to significantly improved exercise performance.

## Sensitivity Analysis

The sensitivity analysis of the disutility of failure in Figure 6 reveals that average personalized utility performance decreases as the disutility effect increases. Interestingly, at lower disutility effect values, personalized utility may slightly decline or even increase. This suggests that the digital health service may lack motivation to improve service quality when users perceive minimal loss from not completing the exercise plans. Similar conclusions were drawn from the experiments conducted with G1, G2 experiment objectives in simulation environments E1, E2, E3, and E4. The digital health service's performance depends on the motivation effect $m$ that drives completion towards the exercise goal, as well as the disutility effect $l$ associated with failing to achieve the exercise goal. These factors can be evaluated through behavioral experiments. Intervention effects, such as gift coupons, can be used to adjust rewards and penalties for goal achievement. Increasing the motivation effect leads to higher average personalized utility performance, with potential differences during skill acquisition and retention stages. In the skill acquisition stage, increasing the motivation effect has a greater impact on exercise performance than in the skill retention stage. One possible reason is that goal motivation significantly encourages the accomplishment of exercise guidance for novices. In addition, when the motivation effect does not reach a threshold, users without digital health service outperform those with digital health service. A possible explanation is that users are more likely to lose interest if the digital health service fails to provide sufficient motivation for daily exercise. Due to the lack of motivation, this can result in outcomes that are inferior to not using the digital health service at all.

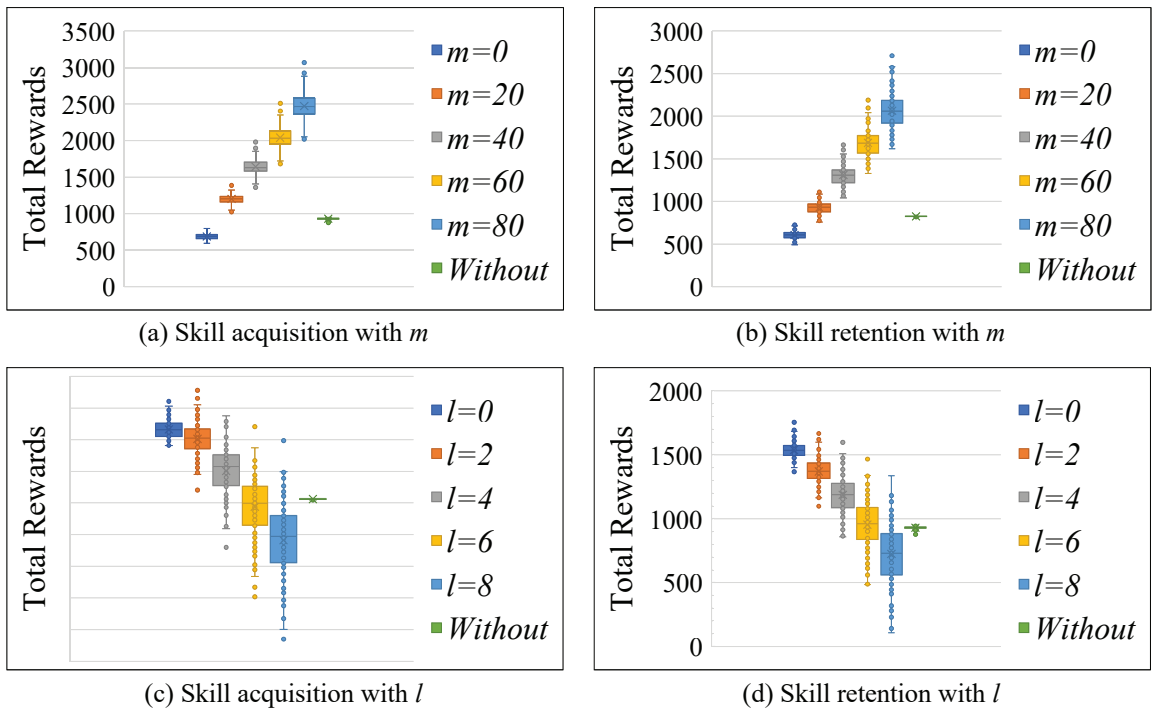

**Figure 6.** Total rewards for different values of $m$, $l$ testing in G3 environment E4.

We further compared our machine learning algorithm with other state-of-the-art algorithms. The competitor algorithms were implemented on the OpenAI-Baseline. For statistical purposes, each simulation was run 100 times. Two of the remaining six other algorithms belong to the same asynchronous advantage actor-critic (A3C) learning algorithm with different network architectures. The network architectures include multi-layer perceptions (MLPs), long short term memory (LSTM) recurrent neural networks.[35] The numbers of neurons in the hidden layers of the A3C-MLPs, A3C-LSTM were determined by empirical experiments: the A3C-MLPs has two hidden layers with 64 neurons, and the A3C-LSTM has a hidden layer with 32 neurons. Others are using different reinforcement learning algorithms with the same network architectures. These hybrid network architectures, which use different reinforcement learning algorithms with exactly the same hybrid network architecture as used in our approach. However, instead of using the A3C reinforcement learning algorithm, these models use different learning algorithms, namely DQN, ACKTR, PPO, and GAIL.[36]

Table 7 shows the results for all algorithms. In terms of computational complexity and average speed, it can be observed that the proposed algorithm consistently outperforms all competitors. Table 7 demonstrates that our proposed algorithm can be executed in 3 seconds during testing. The execution time is close to real time. In addition, our proposed algorithm outperformed other algorithms using the same reinforcement learning algorithm with different network architectures (MLP, LSTM) in averaging speed in the training stage. This indicates that the hybrid neural network architecture is advantageous for learning the user's exercise behavior in the personalized digital health service system. Moreover, our proposed algorithm outperformed others with the same network architecture in different reinforcement learning algorithms (DQN, ACKTR, PPO, and GAIL). This shows the efficiency of the A3C algorithm used to train our hybrid network and find the optimization goal setting strategy for the digital health service.

Table 7. Results of the proposed algorithm and the competitor algorithms (Average value).

| Algorithms | Model | Training Stage | | Testing Stage |
|---|---|---|---|---|
| | | Running Time | Converging Steps | Inference Time |
| **Proposed Algorithm** | Acquisition | 351.09 | 4500 | 2.8 |
| | Retention | 367.46 | 5100 | 2.9 |
| **A3C-MLP** | Acquisition | 426.71 | 5400 | 3.2 |
| | Retention | 410.38 | 5340 | 3.3 |
| **A3C-LSTM** | Acquisition | 431.08 | 4960 | 2.9 |
| | Retention | 405.91 | 5250 | 3.1 |
| **DQN-Hybrid** | Acquisition | 422.13 | 9600 | 3.4 |
| | Retention | 449.12 | 10000 | 3.7 |
| **ACKTR-Hybrid** | Acquisition | 361.61 | 6800 | 3.1 |
| | Retention | 374.27 | 6100 | 2.9 |
| **PPO-Hybrid** | Acquisition | 349.56 | 6400 | 2.9 |
| | Retention | 348.62 | 6260 | 3.2 |
| **GAIL-Hybrid** | Acquisition | 368.43 | 6100 | 3.1 |
| | Retention | 380.19 | 5940 | 2.8 |

In terms of computational complexity and average speed, it can be observed that the proposed algorithm consistently outperforms all competitors. Table 7 demonstrates that our proposed algorithm can be executed in 3 seconds during testing. The execution time is close to real time. In addition, our proposed algorithm outperformed other algorithms using the same reinforcement learning algorithm with different network architectures (MLP, LSTM) in averaging speed in the training stage. This indicates that the hybrid neural network architecture is advantageous for learning the user's exercise behavior in the personalized digital health service system. Moreover, our proposed algorithm outperformed others with the same network architecture in different reinforcement learning algorithms (DQN, ACKTR, PPO, and GAIL). This shows the efficiency of the A3C algorithm used to train our hybrid network and find the optimization goal setting strategy for the digital health service.

## Discussion

### Principal Findings

The experimental results provide valuable information for designing the exercise goal setting strategy in the digital health service. The machine learning approach employed in this study outperformed other fixed exercise goal setting strategies across all assumptions of the user's exercise behavior changes. In addition, it demonstrated the adaptive ability of the machine learning approach to maximize exercise performance during both the skill acquisition and retention stages. Because the experiment showed larger fluctuations and better performance in daily exercise performance compared to not using the digital health service in Figure 4. We visualized the experimental results using stacked column charts (Figure 7). The results indicate that fixed exercise goal setting strategies performed worse than those without digital health service in most cases. In addition, the machine learning method outperformed the strong fixed intensity exercise goal, while the performance improvement was relatively modest for the weak intensity fixed exercise goal and the slightly weak intensity fixed goal. The main reason is that when users encounter obstacles and difficulties during intense exercise goals, it can demotivate them. On the other hand, users who find it challenging to change their exercise behavior may benefit more from setting and achieving less intense exercise goals rather than pushing themselves with high-intensity goals.

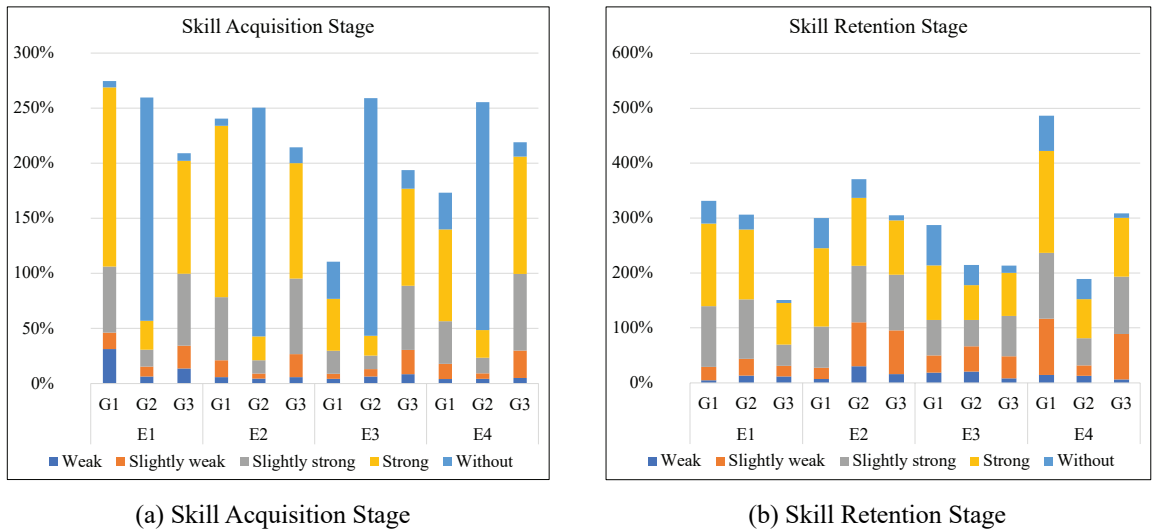

(a) Skill Acquisition Stage  (b) Skill Retention Stage

**Figure 7.** Visualization of the stacked column of the improvement of the machine learning approach over other strategies.

The experiments conducted during the skill acquisition and retention stages provide valuable insights into the service performance of exercise goal setting. We compared the performance of machine learning methods in both stages and visualized the results using clustered column charts, as shown in Figure 8. During the stage of skill acquisition, it is evident that various experimental subjects and users with distinct behavioral tendencies perform differently. However, during the stage of skill maintenance, performance differences are comparatively smaller, indicating more stable performance. These findings indicate that the machine learning approach's continuous analysis and adaptation to changing exercise behavior and user exercise preferences contribute to its efficacy in skill retention. By dynamically adjusting exercise goals, the machine learning approach helps users enhance motivation, and achieve long-term fitness performance.

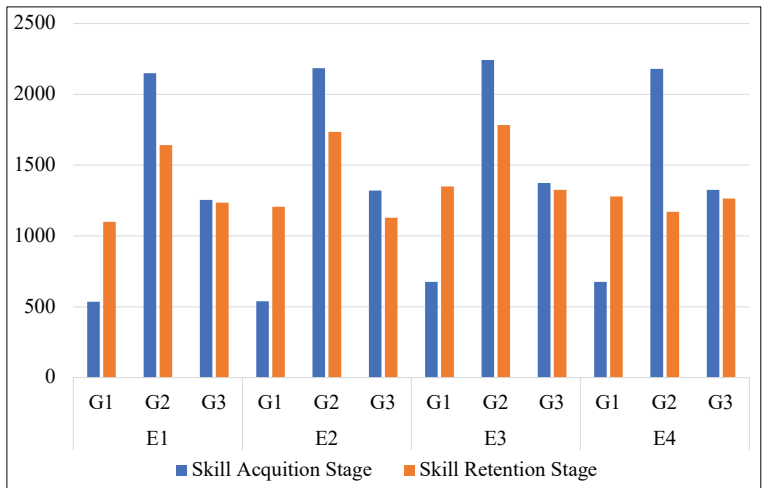

**Figure 8.** Clustered column of total rewards of the machine learning approach.

Through sensitivity analyses, it was discovered that increasing the completion effects results in higher average personalized utility performance, indicating a positive effect of user motivation on exercise goal strategies. When the motivation effect falls below a certain threshold, users without the digital health service performed better than those with the service, indicating that insufficient motivation from the service can lead to user disinterest and potentially worse outcomes than not using the service at all. The analysis of the disutility of failure revealed that as the negative consequences of failure increase, average personalized utility performance decreases, highlighting the need for appropriate incentives and penalties to ensure optimal user motivation.

In digital health service, individual exercise preferences and exercise base level need to be considered when designing exercise goal setting strategies. Different user groups have shown significant variations in exercise performance associated with exercise goal setting strategies, highlighting the necessity of personalized approaches. The adaptability of the machine learning approach to users' exercise behavior and its receptivity to changing preferences and objectives make it a promising option for creating engaging and individualized exercise experiences. Individual preferences and limitations must be considered when designing exercise interventions, as fixed intensity strategies may not be appropriate for all users. These findings provide valuable insights for designing effective exercise interventions in digital health services, highlighting the importance of user motivation, completion effects, and the ineffectiveness of failure in improving performance and overall efficacy.

The implementation of the dynamic exercise goal-setting approach in digital health services benefits users for enhancing user motivation and improving long-term fitness performance by tailoring exercise goals to users' exercise capabilities and modifying them in response to changing behaviors. The dynamic goal setting approach ensures that users engage with achievable exercise goals, resulting in a feeling of fulfilment and enhancing their degree of motivation. The adaptability in the dynamic goal setting approach contributes to the reduction demotivation and the improvement of the overall user experience. Moreover, constant assessment and adjustment of exercise goals in the dynamic goal setting approach improve efficacy during the process of using digital health service, contributing to long-term fitness performance. The dynamic exercise goal setting approach aids in alleviating stagnation and promoting the retention of optimal exercise level during the service period.

## Limitations

A limitation of the exercise goal-setting service in digital health is the absence of measurement for crucial health intervention-related parameter values. This includes both the positive effects of effectively completing the exercise plan and the potential negative effects of not following the plan. This limitation hinders the capability to understand user behavior based on a discrete choice model, necessitating a data-driven model to predict user behavior using retrospective log data. Future research could include behavioral experiments to quantify the effects of the digital health service and determine the optimal reward and punishment levels. This would provide a more comprehensive comprehension of the strategy's effectiveness.

The limited availability of data can introduce biases in the experimental results. Limited availability of data might result in biases in digital health research, as it can be challenging to obtain extensive information on user exercise behaviour. A recent study used simulations and explanatory algorithms to detect true signal in the data instead of perpetuating biases within the training datasets.[9] Therefore, we aim to adopt advanced machine learning models to reduce dataset biases in future research. In addition, the analysis conducted in this study focuses on parameter sensitivity and trend interpretation rather than providing definitive conclusions. These limitations highlight the need for further research and data collection efforts to obtain a more comprehensive understanding of users' exercise behavior and its implications for digital health services. We are going to expand our data collection to involve a more extensive range of users from various backgrounds, therefore obtaining insights on exercise guidance tailored to specific groups, such as for pregnant women with diabetes.[38] On the other hand, recent studies have utilized machine learning methods and factor analysis to identify risk factors in various areas, such as sleep disorders,[10] musculoskeletal complications in diabetes,[39] blood lead levels' impact,[40] and the influence of diet on depression.[41] Drawing on the methods used in these works, we could design machine learning approach to identify the intervention factors associated with exercise performance.

As a methodology study using publicly available datasets, this study avoids testing the algorithm in human experiments due to ethical considerations. Consequently, this study lacks real-time testing data in different types of sensors. A promising direction for future research is testing the algorithm in real situation. We will reinforce data security protocols and establish data encryption protocols in the real environment. At the same time, we will be seeking ethical approval to guarantee adherence to relevant regulations. Hence, we will assess and improve the algorithm in diverse populations and different contexts to ensure the algorithm's efficacy and applicability across various demographic and scenarios in the future.

In the dynamic exercise goal-setting approach, addressing the challenge of long-term user engagement is a crucial area for future research. It is crucial to comprehend the elements that contribute to the perseverance and adherence of users to their exercise routines. One potential strategy for increasing users' adherence to their fitness goals might involve using aspects of gamification, integrating community support, or employing personalized habit-formation strategies. Besides, our exploration of dynamic exercise goal setting approach involves in the impact the mental health in the future, promoting psychological and physical well-being for the users. The potential positive effects on mental health highlighted by recent studies, in particular the correlation observed between engaging in exercise and a reduction in the occurrence of depressive symptoms,[42] suggest that dynamic exercise goal-setting methods might address such data in the future. Furthermore, it has been proven that exercise has therapeutic benefits for the treatment of illnesses including major depression and panic disorder.[43] The insights gained from these studies may in the future help in the development of interventions that are efficient to deal with both physical and mental health problems.

## Comparison with Prior Work

In recent years, there has been a growing focus on improving exercise intervention effects in digital health service. For example, Aswani et al. developed a predictive modeling framework based on qualitative social science models of behavior change in exercise goal setting service.[23] In a similar vein, this paper considers how to improve the service performance for the exercise goal setting strategy for the digital health service.

Our investigation revealed that a diverse range of approaches have been developed for exercise recommendations in digital health service, including mathematical methods and machine learning methods. However, there are limitations associated with each approach. Mathematical methods, which rely on exact models to describe the environment and evaluate exercise performance, lack consideration of human behavioral factors. For instance, Calvert et al. proposed analytical research on exercise performance using a Fitness-fatigue model that predicts exercise performance based on exercise intensity.[17] This model has been applied to predict exercise performance in various endurance sports such as running,[22,44,45] swimming.[17] Nevertheless, in digital health service where the focus is on persuasive information rather than direct control of exercise processes, mathematical models are not directly applicable. On the other hand, machine learning methods, such as artificial neural networks,[46] support vector machine,[47] have been attractive for digital health service, but integrating them with qualitative social science models in exercise goal setting service remains a challenge.[23]

To overcome these limitations, we introduced a state-of-the-art machine learning method, the reinforcement learning method, to enhance the exercise goal setting strategy in digital health service. By incorporating the social science model and adapting to users' heterogeneous characteristics and dynamic health behaviors, the reinforcement learning method offers a promising solution.[13,48] Experimental studies conducted by Forman et al.[48] demonstrated the effectiveness of reinforcement learning in promoting increased exercise and achieving cost savings in digital health service. This research aims to improve exercise intervention effects in digital health service by addressing the limitations of existing approaches. We proposed a machine learning algorithm that learns from prior experiences to predict exercise performance, time-varying health states, and uncertainty regarding future health outcomes. Our algorithm integrated qualitative social science knowledge to improve the interpretability of the deep learning method and provided suggestions on optimal exercise intensity from an exercise process optimization perspective to avoid the potential risks of overexercising. By combining these elements, this study filled a gap in the field and contributed to personalized exercise recommendations in digital health service.

## Conclusions

This paper is an initial exploration of the potential development direction of digital health and offers essential considerations for optimizing and designing self-health management interventions in the future. This research contributes to the growing knowledge in the field of digital health, providing valuable guidance for practitioners and researchers aiming to enhance the design and impact of self-health management services.

# DECLARATIONS

## Conflicts of Interest

The authors declare that they have no known competing financial interests or personal relationships that could have appeared to influence the work reported in this paper.

## Funding

The authors would like to acknowledge the support of the National Natural Science Foundation of China [grant number 72071042].

## Informed Consent

Not applicable – Two publicly available data sets are used in this study is. The first data set is published by Thambawita et al. [33]. The second data set is managed by Open Humans Foundation.

## Ethical Approval

This work did not involve any human or animal trials. The datasets used in this work are anonymous and available in the public domain.

## Guarantor

Haiyan Wang

## Contributorship

The study's conceptualization and conception were done by Ji Fang, Vincent CS Lee, and Hao Ji. Ji Fang conducted literature research, methodology, software development, formal analysis, and data collection, and was responsible for writing the initial draft of the manuscript. Vincent CS Lee supervised the study, provided validation, and contributed to the reviewing and editing of the manuscript. Hao Ji participated in the methodology. Haiyan Wang contributed to project administration, funding acquisition, and providing resources. All authors reviewed and edited the manuscript and approved its final version.

## Acknowledgements

We would like to thank the reviewers for taking the time and effort necessary to review the manuscript.

## Supplementary Material

A. Table: Notations used in this paper. B. Performance evaluation function. C. The formulation of exercise goal setting problem. D. Machine learning approach and algorithm in digital health service. E. Table: Performance comparison with fixed exercise goal setting strategies.